\documentclass[runningheads]{llncs}
\usepackage[T1]{fontenc}
\usepackage{graphicx}

\usepackage{amsmath}
\usepackage{algorithm}
\usepackage{algpseudocode}
\usepackage{float}          %
\usepackage{nicefrac}
\begin{document}
\title{Evolving language compositionality in a frequency-structured meaning space}

\author{Fabio De Ponte\inst{1} \and Eloise Gaines-White\inst{2} \and Conor Houghton\inst{2} \and Seth Bullock\inst{2}}

\authorrunning{F. De Ponte et al.}
\institute{Universit\'{e} de Namur and Vrije Universiteit Brussel, Belgium, 
\email{fabio.de.ponte@vub.be}\\
\and
University of Bristol, UK\\
\email{\{bq23138,conor.houghton,seth.bullock\}@bristol.ac.uk}}
\maketitle              %
\begin{abstract}
 The iterated learning model was introduced to investigate language evolution: the way in which the characteristic properties of human languages have been shaped, at least partly, by repeated transmission from one language user to another. The key finding is that language compositionality can arise spontaneously as a consequence of language being passed repeatedly through a language learning bottleneck. Here we explore how changing the frequency of different meanings, so that some meanings occur much more frequently than others, affects the character of its compositionality. We find that, as observed in natural languages, high-frequency meanings can escape the pressure to conform to the grammar that characterizes lower-frequency meanings. However, when the frequency structure is instead imposed on parts rather than on whole meaning vectors, the language fails to transmit across generations. This occurs despite the fact that the most frequent elements are reliably learned. These results suggest that frequency can shape emergent linguistic structure only when the frequency distribution is defined over form-meaning units that learners can acquire holistically. When frequency is instead distributed over smaller units, it fails to support the relational structure required for compositional generalisation, thereby preventing stable language transmission.

\end{abstract}

\section{Introduction}
Language evolution has been studied from a range of perspectives, including approaches that conceptualise language as a complex adaptive system \cite{Wacewicz2021,kirby2011language}. This perspective has informed both theoretical and empirical research, including usage-based frameworks such as construction grammars \cite{tomasello2003constructing,beuls2023construction} and computational approaches such as iterated learning models (ILMs), which provide a framework for studying the evolution of language properties across generations.

In this paper, we use the semi-supervised ILM version \cite{bunyan2025iterated,bullock2024modeling}, which was developed to address limitations of the original model \cite{kirby2000syntax,kirby2001spontaneous,kirby2002learning,kirby2002emergence}. In particular, the original ILM includes a step of ``obversion'', in which the learned dictionary is reversed from meaning-to-signal to signal-to-meaning. This operation is both computationally demanding and unrealistic as a simulation of human language learning. The semi-supervised ILM avoids this reversal step, providing a more computationally tractable and cognitively plausible formulation. The model has been used to investigate in particular three properties of language: \emph{expressivity}, \emph{compositionality}, and \emph{stability}.

Expressivity refers to the extent to which a linguistic system can distinguish between meanings, or, equivalently, the degree of ambiguity in the language. Natural languages achieve remarkable expressivity, allowing speakers to convey highly unusual meanings or meanings that may never have been expressed before. In our model, the measure is operationalised as the proportion of meanings that can be expressed through distinct signals. 

Compositionality refers to the extent to which the mapping between meanings and signals systematically reuses parts of signals to express parts of meanings. For example, the concept `to research' is consistently expressed by the word `research' across a range of English utterances and contexts, such as `I research', `we researched', `they are researching', `a research paper', and `a job in research'. A defining property of compositionality is therefore that small changes in meaning tend to correspond to small changes in the associated signal. An interesting feature of natural languages is that the \emph{regularity} of this reuse can vary with the frequency of the meaning. For example, unlike the relatively rare verb `to research', which follows a highly regular pattern, the very common meaning `to be' is expressed by forms that differ substantially in English, such as `I am', `we were', and `they are'.

Stability is a measure of the similarity between the language taught by the tutor and the language learnt by the pupil. It is calculated as the proportion of meanings for which the same signal is used by both tutor and pupil. 

In this paper, we further extend the model by investigating how the frequency structure of the meaning space influences the evolution of the three properties discussed above. In particular, we examine whether assigning different frequencies to meanings affects the extent to which these properties emerge and are maintained across generations. Our analysis focuses primarily on \emph{compositionality}, asking whether the distribution of meaning frequencies creates selective pressures that favour or constrain the emergence of systematic relationships between meaning components and form components.

\subsection{The semi-supervised iterated learning model} 
The iterated learning model (ILM) simulates a scenario in which an initially naive pupil acquires language from a tutor and subsequently becomes the tutor for a new naive pupil in the next generation. At each transmission step, language passes through a bottleneck \cite{kirby2002learning,brace2015achieving,lind2025sequence}: the pupil is not exposed to the entire language but only sees a subset of meaning-signal pairs. As a consequence, the learner is required to generalise to the unseen portion of the language.

In the semi-supervised ILM used here \cite{bunyan2025iterated}, each agent is modelled as a neural network, specifically an autoencoder consisting of an encoder and a decoder. The encoder maps meanings to signals, a process we refer to as \textit{encoding}, while the decoder performs the inverse operation by reconstructing meanings from signals, a process we refer to as \textit{decoding}. In each case, a particular meaning takes the form of a bitstring and the signal that expresses this meaning in a particular language is also a bitstring of the same length. An agent's language is thus the mapping from each possible meaning bitstring to the signal bitstring that expresses it.

At the start of each generation, a naive pupil is created with no knowledge of language. Their tutor then generates a set of utterances in their own language, pairing each meaning in a randomly sampled subset of the meaning space with the associated signal used to express this meaning in the tutor's language. The pupil's encoder and decoder are then trained on this subset: for the encoder, meanings serve as inputs and signals as targets, while the reverse is true for the decoder. Each batch contains a single sample to train the encoder and a single sample to train the decoder, together with a larger number of meanings used to further train the pupil's whole autoencoder, used as both inputs and targets.

\section{The interpretation} 

Compositionality is a property that operates across multiple levels of linguistic structure. A sentence such as ``The brown dog is in the garden'' describes a particular situation that can be modified through the replacement of individual words. For example, changing dog to cat produces ``The brown cat is in the garden'', while replacing garden with kitchen yields ``The brown dog is in the kitchen''. In both cases, a local modification produces a limited and systematic change in meaning. By contrast, replacing several words simultaneously can lead to a substantially different, but still systematic, interpretation, as in ``A blue car is in the garage''. The same principle applies below the word level. Words such as ``actor'' and ``action'' share the root ``act'', which contributes a core semantic component, while suffixes such as -or, -ion encode additional information. Arguably, compositionality also extends to higher levels of discourse, where replacing or modifying entire paragraphs can alter the overall meaning and communicative effect of a text in a systematic manner.

The semi-supervised ILM investigates the mechanisms underlying compositionality and is not restricted to any particular level of linguistic organisation. Consequently, bitstrings and individual bits may be interpreted as corresponding to different representational layers. In this paper, for ease of exposition, we treat each signal bitstring as an object and individual bits as its properties. Under this interpretation, an entire meaning bitstring represents a situational setting or scenario that must be communicated, while the corresponding signal bitstring is interpreted as a holophrastic utterance.

Constructivist accounts of language acquisition propose that linguistic knowledge emerges gradually through children's exposure to, and participation in, language use \cite{macwhinney1975rules,macwhinney1982basic,peters1983units,pine1997slot,tomasello2003constructing,tomasello2009usage,ambridge2015constructivist}, rather than being determined by a pre-specified biological template. The process relies on pattern recognition: the ability to identify similarities and differences across sensory and motor experiences and to use these regularities to construct categories and cognitive schemas. More specifically, during early language development, children begin establishing broad associations between the linguistic forms they hear and the situational contexts to which these forms refer. These associations are known as holophrase constructions, because children initially process them as indivisible wholes rather than analysing them into smaller meaningful components. Holophrases may correspond to multiword expressions that children interpret as single units, such as ``Look at the black dog''. Subsequently, exposure to expressions such as ``Look at the black cat'', referring to a similar situation except for the substitution of the animal, allows children to infer that a subpart of the utterance, such as ``dog'', corresponds to a specific component of the situation and can be systematically replaced. This process marks the beginning of the emergence of compositional structure.

The usage-based approach has led to major advances in the computational modelling of language learning within a constructionist framework, particularly through the Fluid Construction Grammar paradigm \cite{beuls2023fluid,beuls2024humans,beuls2023construction}. This line of research has demonstrated that large-scale, human-interpretable construction grammars can be learned directly from language use and can capture complex syntactic and semantic regularities \cite{van2026method}. While successful in modelling acquisition from existing linguistic data, this approach does not address the broader evolutionary question of how language itself changes over time or how linguistic systems emerged in the first place \cite{cangelosi2002computer}. This is the focus of the ILM \cite{van2011can}. 

As mentioned in the introduction, language is defined as a mapping between meaning vectors and signal vectors. An individual bit $i$ in a meaning vector $\mathbf{m} \in M$ will be referred to as a ``fact''. Similarly, a bit $j$ in a signal vector $\mathbf{s} \in S$ will be referred to as a ``word''. Strictly speaking, ``morpheme'' would be a more accurate term, since the signal vector is intended to model any aspect of how meaning is encoded. However, for convenience, we use ``word'' throughout.

In this paper, we interpret meaning bitstrings as representations of situational settings. For example, consider a set of objects that vary in terms of an ordered set of binary properties such as ``large'' versus ``small'' or ``black'' versus ``white''. Each bit of a meaning bitstring might encode one of these properties. With an 8-bit meaning representation, we can map $2^8$ scenarios: the first bit may indicate size, the second colour and so on. Likewise, a specific combination of signal bits represents a holophrastic sentence used to convey a specific situational setting, in this case, a particular meaning. For the first tutor, signal bitstrings are arbitrary with respect to meanings and tend to exhibit no predefined internal structure. Nothing prevents multiple meanings from being mapped to the same signal and, equivalently, there is no guarantee that every signal is employed within a mapping. Importantly, there is no guarantee that an arbitrary mapping between meanings and signals can be learned from exposure to a subset of the mapping, because it is likely that there is no information in one part of the mapping that can help to anticipate the structure of some other part.

\section{Methods} 

\subsection{The experiment} 

At each generation, a subset $B$ of meanings is selected and used to create a set of meaning-signal pairs. It is then used for supervised training: the encoder is trained on $B_1 = B$, and the decoder on $B_2$, a shuffled copy of $B_1$. Concurrently, a subset $A$ of meanings, typically larger than $B$, is sampled from the full meaning space to train the autoencoder in an unsupervised fashion. Within each epoch, the encoder and decoder each receive $|B|$ gradient updates. For every iteration, $r$ additional autoencoder updates are performed on meanings drawn independently from $A$. This yields a $1{:}1{:}r$ update ratio between encoder, decoder, and autoencoder, so the autoencoder receives $r \times |B|$ gradient steps per epoch compared to $|B|$ for the supervised components.

In \cite{bunyan2025iterated} it is demonstrated that this experimental setup can achieve optimal compositionality, expressivity, and stability. Figure~\ref{flat_vs_zipf_meaning}a shows the evolution of all three metrics for a 16-bit encoder-decoder model over 50 generations, under the configuration $|A| = 480$, $|B| = 320$, learning rate $0.5$, and 20 autoencoder training iterations for every encoder and decoder learning iteration, that is a loop ratio of $1{:}1{:}20$. Results are averaged over 30 independent trials. Under these conditions, all three metrics converge to their maximum value within approximately 20 generations.

Here, we extend this by investigating how language compositionality is shaped when learners are exposed to an uneven distribution of meanings. Specifically, we address two questions: first, whether meaning vectors, (representing situational settings) that occur more frequently during training lead to signals that are more or less compositional; and second, what effects emerge when the uneven meaning distribution applies at the level of individual bits rather than at the level of whole meanings.

To address these questions, we impose an uneven frequency profile on meanings, first \textit{within the whole meaning vector space} and then \textit{between the individual bits} of the meaning vectors. We use a Zipf distribution to model this frequency profile, reflecting its well-established role in the frequency structure of natural language \cite{george1949zipf,saichev2009theory,wyllys1981empirical}. Zipf's law \cite{aitchison2016zipf} states that the frequency of each element of the set is inversely proportional to its rank: $f(r) \propto 1/r^{\alpha}$, where $r$ is the rank, the position of the element in a frequency ordered list, and $\alpha$ is an exponent controlling the steepness of the distribution. Setting $\alpha = 0$ recovers a uniform distribution, in which all meanings are equally frequent, whereas larger values of $\alpha$ concentrate probability mass on the highest-ranked meanings. By sampling meanings according to this distribution during training, we expose the learner to a skewed input in which a few meanings recur frequently and a long tail of meanings are encountered only occasionally or not at all, allowing us to isolate the effect of frequency on the emergence of compositional structure.

To apply a Zipf distribution over the set of $n$-bit meaning vectors, the $2^{n}$ meanings are first shuffled and then each is given a probability proportional to $1/\mathrm{rank}^{\,\alpha}$, where rank is the new rank position of the meaning. These probabilities are preserved from generation to generation.

We also consider a different manipulation, in which the Zipf distribution is applied to the individual bits within each meaning vector. During training, each meaning vector is generated independently by sampling a bit at each position according to its own probability of taking the value 1, with these probabilities following a Zipf distribution across bit positions. For example, with $\alpha = 1$, the probability that the second highest ranking bit is set to 1 is half the probability that the highest ranking bit is set to 1.

Finally, we consider an intermediate manipulation in which the Zipf distribution is applied to contiguous chunks of the meaning vector, rather than to whole meanings or to individual bits. Chunk size determines the granularity of the frequency bias: a chunk size of $n$ corresponds to sampling complete meaning vectors, while a chunk size of 1 applies the distribution independently to each bit. Intermediate chunk sizes interpolate between these two extremes. Chunk-level sampling also helps clarify the difference between meaning-vector-level sampling and bit-level sampling. At first sight, applying the skewed distribution to individual bits and applying it to chunks of size 1 appear equivalent. However, they are very different. When sampling at the bit level with $n=16$ and $\alpha=1$, the highest-ranked bit has probability $0.5$ of being $1$, the second-ranked bit has probability $0.25$, the third-ranked bit has probability $\approx 0.17$, and so on. In contrast, when the skew is applied to chunks, the ranking is defined within each chunk, not between them. For chunks of size 1, each chunk can only take two possible states, causing the distribution to collapse into a binary bias where each bit has probability $0.66$ of being $1$ and $0.33$ of being $0$. This affects in a different way the distribution of meaning vectors.

\subsection{Metrics} 

To measure the progress of language learning at each generation, we will use three metrics: expressivity, compositionality and stability.

\textbf{Expressivity} quantifies the diversity of the signals produced by the pupil after learning, and hence the lack of ambiguity in their language. Specifically, it is the number of distinct signals employed in the language normalised by the number of possible signals available, which is $2^n$ if signals are bitstrings of length $n$. A fully expressive language uses every possible signal exactly once; a degenerate one reuses the same signal to encode multiple meanings.

\textbf{Compositionality} quantifies the degree to which individual bits of a signal (the words) systematically correspond to individual bits (the facts) of meaning. The metric is computed by identifying, for each meaning bit, the signal bit that provides the most reliable information about it, i.e., the one associated with the lowest conditional entropy. Entropy, for a probability $p$, is defined as:
\begin{equation}
    H(p) =
    \begin{cases}
        -p \log_2(p) - (1 - p) \log_2(1 - p) & \text{if } p \in (0, 1) \\
        0 & \text{otherwise}
    \end{cases}
    \label{eq:entropy}
\end{equation}
Where multiple meaning bits are best encoded by the same signal bit, the conflict is resolved greedily: the best-fitting pair retains its entropy score, while all remaining competing bits are penalised with an assigned entropy of $1.0$. The final compositionality score is then $1 - \bar{H}$, where $\bar{H}$ is the mean adjusted entropy across all meaning bits.

From this procedure, two complementary compositionality metrics are calculated. The first operates at the vector level: it measures the extent to which a given meaning vector is encoded by a signal vector that conforms to the global grammar defined through the process above. The second operates at the bit level: it measures the extent to which a single signal bit systematically encodes any single meaning bit across the entire language. The pseudocode for these measures is presented in the Appendix.

\textbf{Stability} measures the extent to which the pupil's language matches the tutor's language after learning. As with compositionality, two complementary metrics are calculated. The first operates at the vector level: at each generation, it is defined as the proportion of meanings, across the entire meaning space, for which the pupil exactly reproduces the tutor's encoding, providing a snapshot of how faithfully the language as a whole is transmitted (Figure~\ref{flat_vs_zipf_meaning}, left panels, and Figure~\ref{2x2_1_0_bit}, panel \textit{a}). The second operates at the bit level: for each meaning bit position, the bit stability score is the fraction of generations, over the whole chain, in which that position is encoded by the same carrier signal bit, identified as its most frequent one; low scores thus indicate a meaning bit that frequently switched between carriers (Figure~\ref{bit_stability_compositionality_by_rank}, right panel).

As a baseline, we run the model with $\alpha = 0$, corresponding to a uniform frequency distribution over meanings during training. Figure~\ref{flat_vs_zipf_meaning} shows that, under this condition, compositionality, expressivity and stability all rapidly converge to their maximum values and sentence-level compositionality rises uniformly for all sentences, reproducing previously reported results \cite{kirby2002learning,bunyan2025iterated}.

\section{Results} 

\begin{figure}

\includegraphics[width=\textwidth, trim=1pt 1pt 1pt 1pt, clip]{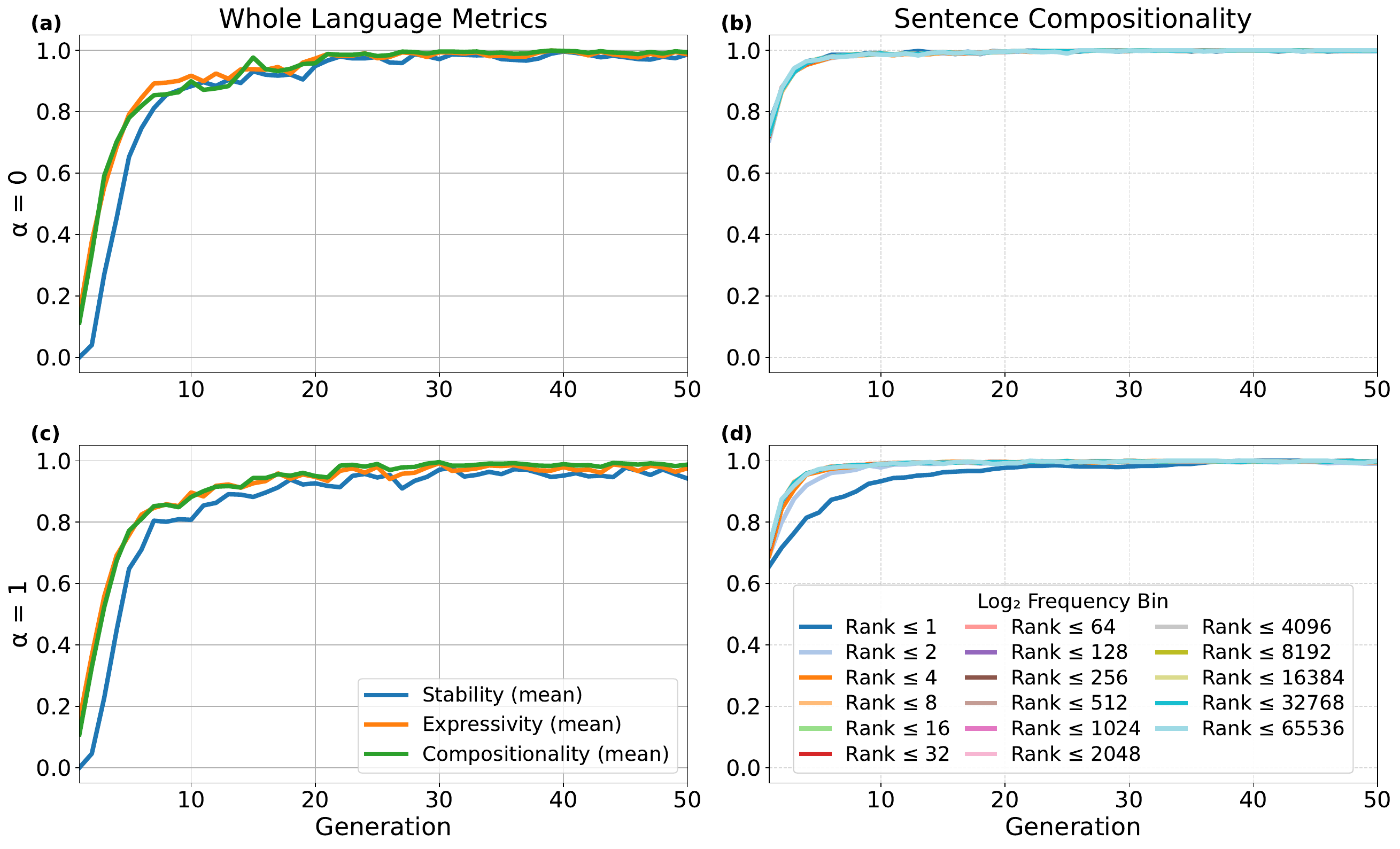}

\caption{\textbf{Flat vs.\ whole-meaning Zipf distribution:} Language evolution in the semi-supervised iterated learning model for a 16-bit encoder-decoder neural network over 30 trials of 50 generations, under the configuration $|A| = 480$, $|B| = 320$, loop ratio $1{:}1{:}20$, and learning rate $0.5$. Results are shown for a flat distribution ($\alpha=0$, upper two panels) and a Zipf distribution over meaning vectors ($\alpha=1$, lower two panels). All results are averaged over 30 independent trials. The two left-hand panels show compositionality, expressivity and stability over generations. The two right-hand panels show per-meaning compositionality grouped (averaged) into $\log_2$ bins of rank. The first bins (high-rank) correspond to the most frequent meanings, while later bins (low-ranks) correspond to increasingly rare meanings. For $\alpha=0$ all meanings are sampled uniformly, so the bins are equivalent and the curves coincide, rising rapidly to near 1. For $\alpha=1$, the compositionality of signals conveying the most frequently communicated meanings rise more slowly, only approaching one much later in language evolution.}
\label{flat_vs_zipf_meaning}
\end{figure}

Figure~\ref{flat_vs_zipf_meaning}c shows that, when a pupil is exposed to an uneven distribution of samples from the tutor's language, a compositional language arises and remains stable for the rest of the chain: at the aggregate level compositionality, expressivity and stability all climb within the first 10 generations and then plateau near~1 around generation 30. However, Figure~\ref{flat_vs_zipf_meaning}d shows that the most frequently communicated meanings (for example, rank~$\leq 1$, dark blue line) achieve compositionality more slowly than those that are communicated more rarely. Meanings that occur more frequently tend to be associated with holophrastic sentences, while meanings encountered less often must be encoded compositionally via systematic regularities within the language.

\begin{figure}
\includegraphics[width=\textwidth, trim=1pt 1pt 1pt 1pt, clip]{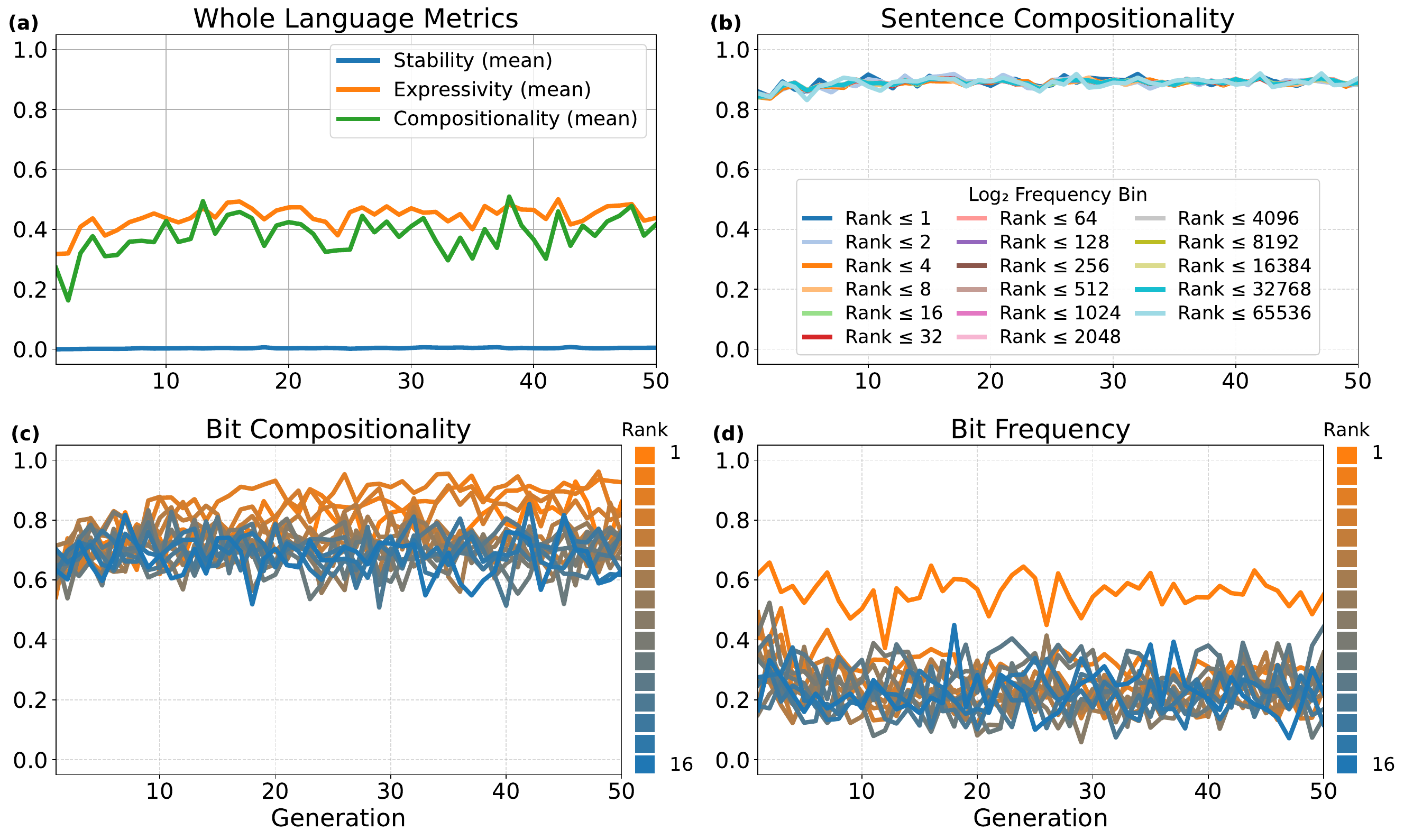}
\caption{\textbf{Bit-level Zipf distribution:} Language evolution under bit-level Zipf distribution ($\alpha=1$), where the frequency skew is applied between bit positions rather than within the meaning vector space. In all other respects the simulations follow the same scheme and parameters as those depicted in Figure~\ref{flat_vs_zipf_meaning}. In panels~\textit{c} and \textit{d}, the signal bits are ranked by how often they are set to 1 across all signals used in the pupil's language.}
\label{2x2_1_0_bit}
\end{figure}

\begin{figure}
\includegraphics[width=\textwidth, trim=1pt 1pt 1pt 1pt, clip]{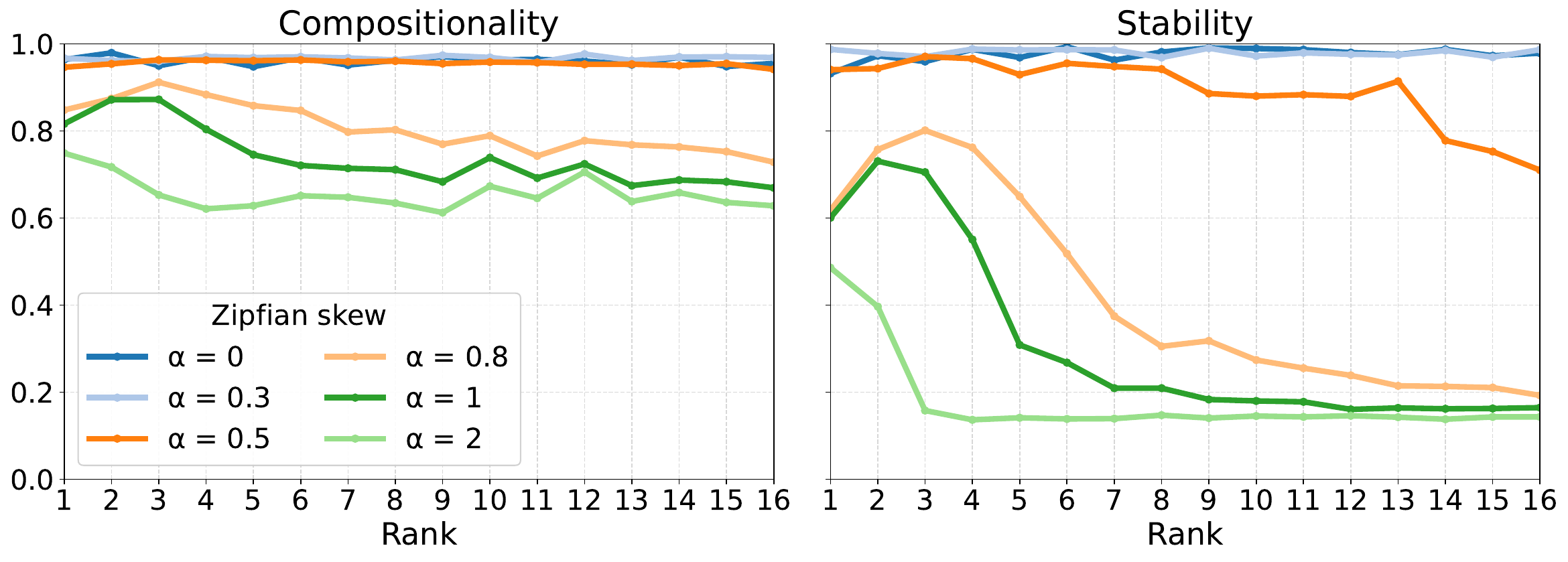}
\caption{\textbf{Per-bit compositionality and stability by frequency rank:} the
highest-ranked meaning bits are associated (left panel) with the most compositional signal bits. The four highest-ranked bits exhibit higher compositionality than bits ranked fifth and below, although rank 1 shows slightly lower compositionality than rank 2. The highest-ranked bits exhibit substantially greater stability (right panel) than lower-ranked ones.}
\label{bit_stability_compositionality_by_rank}
\end{figure}

Then we turn to what happens when the uneven distribution is applied to the bit level: the results with $\alpha=1$ are shown in Figure~\ref{2x2_1_0_bit}. Panel~\textit{a} shows that the chain fails to converge: stability remains close to zero across generations, while expressivity and compositionality fluctuate around 0.4 without reaching a stable regime. Because the Zipf distribution is applied at the level of bit positions, the per-sentence analysis (panel~\textit{b}) reveals no clear structure. However, the per-bit frequency analysis uncovers a pronounced gradient (panel~\textit{d}); higher-ranked meaning bits tend to exhibit greater compositionality, as illustrated in panel~\textit{c} (per generation) and in Figure~\ref{bit_stability_compositionality_by_rank} (per rank). Finally, although the language as a whole does not reach stability, the highest-ranked bits exhibit substantially greater stability than lower-ranked ones.

What appears to be happening is that pupils consistently fail to reproduce the exact signals used by their tutors, so language stability remains close to zero. Nevertheless, the chain of independently emerging languages do share some common internal structure. In particular, the highest-ranked meaning bits (1--3) are consistently mapped to signal bits with high compositionality, indicating that these signal bits reliably track the corresponding meaning bits. These signal bits therefore carry the greatest amount of information. By contrast, lower-ranked meaning bits are mapped to signal bits with substantially lower compositionality. This is likely because, under the Zipfian distribution, low-ranked meaning bits are almost always observed as `0' during training, providing little pressure for learners to encode them systematically. Although the identity of the high-ranking meaning bits varies randomly across different simulations, this pattern is remarkably consistent across independently evolved languages. The signal bit that is most frequently set to `1' typically takes this value about 50\% of the time and also exhibits the highest compositionality, while progressively rarer signal bits tend to exhibit progressively lower compositionality.

Finally, we look at what happens when the uneven distribution is applied at the chunk-level. In Figure~\ref{sentence_compositionality_by_rank} we can see that, as expected, with a chunk size of 16, the effect almost perfectly overlaps with sentence-level sampling: the first bin exhibits the lowest compositionality, while the remainder of the curve tends to remain flat. With a chunk size of 8, compositionality initially remains flat and then gradually decreases with rank. In contrast, when chunks have size 1, 2, or 4, compositionality shows a more pronounced decrease with rank.

\section{Discussion} 

Our results show that the effect of frequency on emergent compositionality depends critically on \emph{what} the frequency structure is imposed upon. When a Zipf distribution is placed over whole meanings the model reproduces a well-known property of natural language: the most frequent meanings converge more slowly to the systematic grammar and remain less compositional than rarer ones (Figure~\ref{flat_vs_zipf_meaning}d). This suggests that a meaning that recurs in nearly every training batch is encountered often enough to be fixed as an idiosyncratic, holophrastic whole, and is under less pressure to conform to the regularities that organise the rest of the language (like the irregular conjugation of the very frequent verb ``to be''). A rare meaning, by contrast, is seen too seldom to be memorised individually; the only way it can survive repeated transmission through the bottleneck is by inheriting and thereby conforming to the systematic meaning--signal mapping (like the regular conjugation of the infrequent verb ``to research'').

\begin{figure}
\includegraphics[width=\textwidth, trim=1pt 1pt 1pt 1pt, clip]{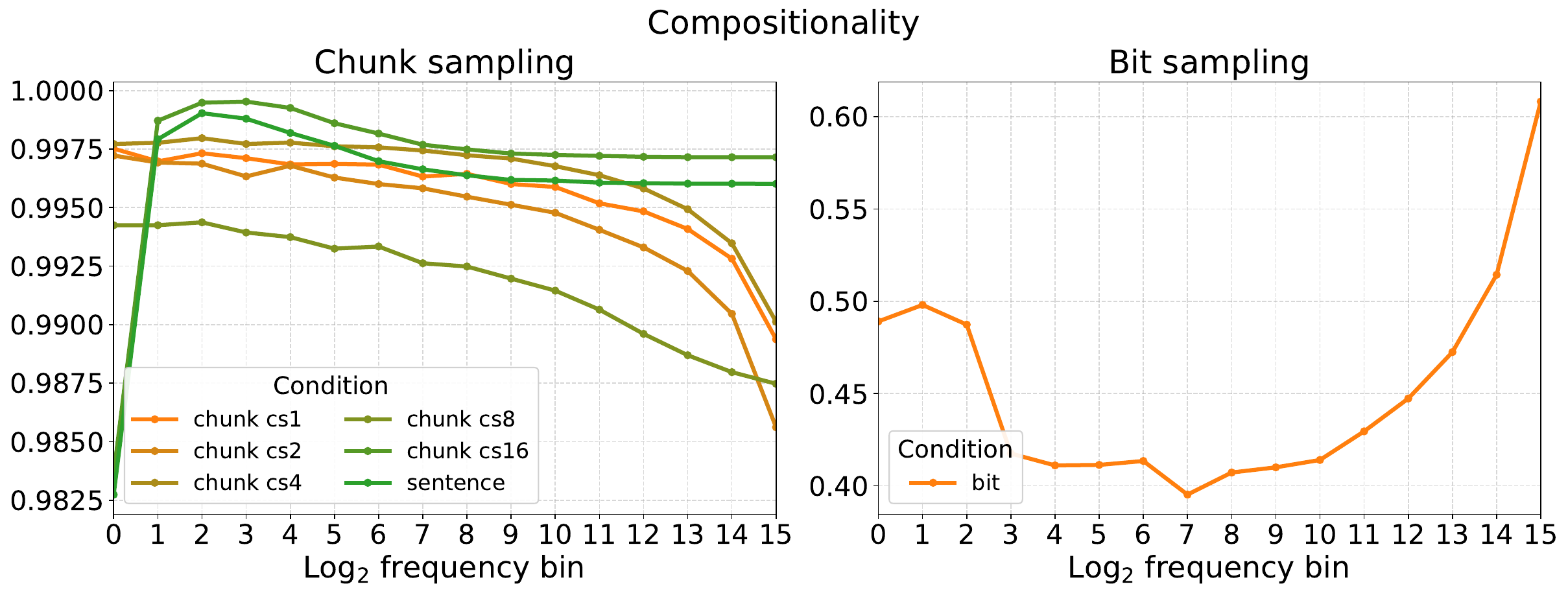}
\caption{\textbf{Compositionality by $\log_2$ frequency-rank bin:} meanings are grouped into logarithmic frequency-rank bins, with bin~0 containing the highest-frequency meaning, bin~1 the next highest-frequency group, and so on. With a chunk size of 16, the effect almost perfectly matches whole-vector-level sampling: the first bin displays the lowest compositionality, while the rest of the curve remains approximately flat. For a chunk size of 8, compositionality is initially stable but gradually declines with increasing rank. By contrast, when chunks are smaller (sizes 1, 2, or 4), compositionality exhibits a decrease across rank.}
\label{sentence_compositionality_by_rank}
\end{figure}

Chunk-level sampling clarifies that this effect is a property of skew over whole meanings, rather than of skew \emph{per se}. As shown in Figure~\ref{sentence_compositionality_by_rank}, with a chunk size of 16, the signal associated with the highest-frequency meaning vector exhibits the lowest compositionality. In contrast, with a chunk size of 1, signals associated with the highest-frequency meaning vectors are the most compositional. Moreover, under bit-level sampling (right panel), emergent languages exhibit a qualitatively different pattern: overall compositionality is substantially lower, and lower-frequency meanings are mapped to signals with higher compositionality.

Moreover, when a Zipf distribution is imposed over meaning bits, no transmissible language emerges. Aggregate stability remains close to zero across generations (Fig.~\ref{2x2_1_0_bit}, panel~\textit{a}), indicating that no shared mapping is consistently transmitted over time. Nevertheless, high-rank meaning bit positions (those most frequently taking the value 1) become relatively stably associated with signal bits and achieve higher levels of per-bit compositionality, whereas low-rank positions switch more often between carriers and remain only weakly systematic (Fig.~\ref{bit_stability_compositionality_by_rank}). As a result, the language develops locally stable and compositional regularities without ever consolidating into a globally stable system.

\section{Conclusion}

We use a semi-supervised iterated learning model to investigate how imposing frequency structure on a meaning space shapes the compositionality of emergent languages. Our central finding is that the effect of frequency skewness does not depend on its presence \emph{per se}, but on the level of representation at which it is introduced. When entire configurations are presented frequently, they tend to be memorised as fixed units and associated with holophrastic phrases. In contrast, when individual elements occur frequently across different configurations, independently of their context, they become strongly associated with lexical forms but lose their role within a broader structural system. The consequence is that no stable language is transmitted across generations. Although the high-rank meaning bit positions may become locally more compositional by acting as reliable carriers, low-rank positions remain largely uninformative. The resulting system lacks the relational structure required for compositional generalisation.

These findings are consistent with usage-based accounts of language, in which grammatical knowledge emerges from recurring constructions rather than from isolated symbol--referent associations. A frequent element learned independently may acquire a stable form, but successful language transmission requires preserving the relationships that bind elements into larger meaningful configurations. Frequency therefore supports the emergence of structure only when it shapes units that learners can acquire holistically; when applied to isolated sub-symbolic features, it fails to provide the organisation needed for generalisation beyond limited exposure.

\section{Further work}

In this work, we investigated how differences in exposure to meaning-signal pairs during language acquisition affect the evolution of language across generations, and how compositionality can emerge from this process.

Our experiments were, however, limited to settings in which the length of the signal bitstring is equal to the length of the meaning bitstring. A possible direction for future work is therefore to extend the model to settings in which the two have different lengths and to adapt the metrics accordingly.

Another possible line of research is to investigate how the compositional mechanisms observed in our model could develop into the hierarchical structures characteristic of natural languages. This could be explored by moving beyond frequency distributions and introducing \emph{dependencies} among meaning components. In the experiments presented here, each bit of a meaning vector was sampled independently. In natural communication, however, meaning components are not independent but tend to co-occur in structured ways: for example, only certain entities are likely to be hairy, and only some are likely to have tails that wag.

We therefore propose constraining the generation of meanings during training by introducing dependencies of this kind. A bit may take the value~1 only when a set of dependent bits satisfies a specific configuration. The key question is whether the ILM will favour languages whose compositional structure captures these dependencies, such that a hierarchically organised meaning space gives rise to a hierarchically organised language purely through transmission across a learning bottleneck.

\bibliographystyle{splncs04}
\bibliography{bibliography}

\section{Appendix}

The pseudocode below describes three compositionality measures that capture complementary aspects of language structure. The bit-level metric identifies the most reliable one-to-one correspondences between meaning bits and signal bits by measuring the consistency of each possible mapping through conditional entropy, thereby extracting a global bit-level grammar. The vector-level metric evaluates how well individual meaning vectors conform to this grammar by measuring the proportion of bit-level rules correctly expressed by their corresponding signals. Finally, the whole-language metric aggregates these local correspondences into a single global score, quantifying the extent to which the entire language can be described by a consistent compositional encoding.

\begin{algorithm}
\caption{Bit-Level Compositionality}
\label{bitlevelcompositionality}
\begin{algorithmic}[1]
\For{each pair (meaningBit $i$, signalBit $j$)}
\Comment{Two-sided binary entropy function}
    \State $p_1 \gets P(\text{signal}_j = 1 \mid \text{meaning}_i = 1)$
    \State $p_0 \gets P(\text{signal}_j = 1 \mid \text{meaning}_i = 0)$
    \State $\text{entropy}(i,j) \gets \frac{1}{2} H(p_1) + \frac{1}{2} H(p_0)$
\EndFor
\State
\Comment{Greedy one-to-one assignment across whole language}
\While{unassigned bits remain}
    \State Pick $(i^*, j^*)$ with lowest $\text{entropy}(i,j)$
    \State Mark $i^*$ and $j^*$ as used
    \State $\text{compositionality}_{j^*} \gets 1 - \text{entropy}(i^*, j^*)$
    \State $\text{correlation}_{j^*} \gets \text{sign}(2p_1 - 1)$
\EndWhile
\end{algorithmic}
\end{algorithm}

\begin{algorithm}[H]
\caption{Vector-Level Compositionality}
\label{sentencelevelcompositionality}
\begin{algorithmic}[1]
\Require Bit-level grammar (Algorithm~\ref{bitlevelcompositionality}) already computed
\For{each meaning $c$}
    \State $\text{rank} \gets \text{Zipf rank of } c$ \Comment{from shuffled permutation}
    \State $\text{bin} \gets \lfloor \log_2(\text{rank} + 1) \rfloor$
    \State
    \For{each (signalBit $j$, meaningBit $i$) pair in the grammar}
        \State $\text{expected} \gets \text{meaning}_i[c]$
        \State count whether $\text{pupilSignal}_j[c]$ matches $\text{expected}$
    \EndFor
    \State $\text{comp}[c] \gets$ fraction of rules that matched
    \State
\EndFor
\end{algorithmic}
\end{algorithm}

\begin{algorithm}
\caption{Whole-Language Compositionality}
\label{wholelevelcompositionality}
\begin{algorithmic}[1]
\For{each pair (meaningBit $i$, signalBit $j$)}
    \State $p_1 \gets P(\text{signal}_j = 1 \mid \text{meaning}_i = 1)$
    \State $p_0 \gets P(\text{signal}_j = 1 \mid \text{meaning}_i = 0)$
    \State $\text{entropy}(i,j) \gets \frac{1}{2} H(p_1) + \frac{1}{2} H(p_0)$
\EndFor
\State
\Comment{Greedy assignment with collision resolution}
\For{each meaningBit $i$}
    \State $j^* \gets \arg\min_j \ \text{entropy}(i, j)$
    \If{$j^*$ already claimed by a better-fit meaningBit}
        \State $\text{adjustedEntropy}[i] \gets 1.0$ \Comment{penalty}
    \Else
        \State $\text{adjustedEntropy}[i] \gets \text{entropy}(i, j^*)$
    \EndIf
\EndFor
\State
\State $\text{compositionality} \gets 1 - \dfrac{1}{n}\sum_{i} \text{adjustedEntropy}[i]$
\State \Return $\text{compositionality}$
\end{algorithmic}
\end{algorithm}

\begin{figure}
\includegraphics[width=\textwidth, trim=1pt 1pt 1pt 1pt, clip]{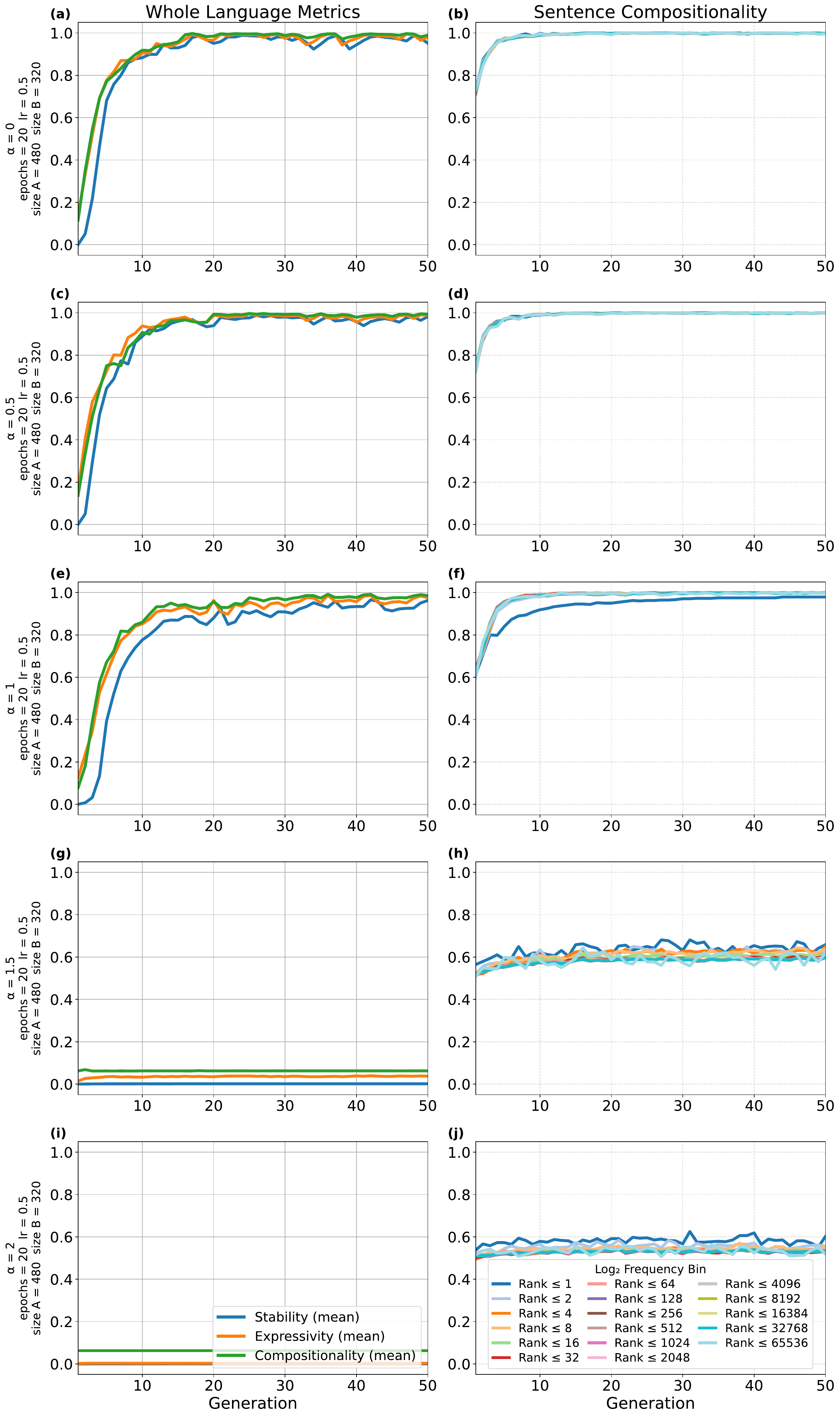}
\caption{Distribution skewed at the meaning-vector level. Left: expressivity, compositionality, and stability for different values of $\alpha$. Right: compositionality as a function of $\log_2$ frequency-rank bins. Meanings are grouped into logarithmic frequency-rank bins, with bin~0 containing the highest-frequency meaning, bin~1 the next highest-frequency group, and so on.}
\label{appendix01_all_alpha_comparison}
\end{figure}

\begin{figure}
\includegraphics[width=\textwidth, trim=1pt 1pt 1pt 1pt, clip]{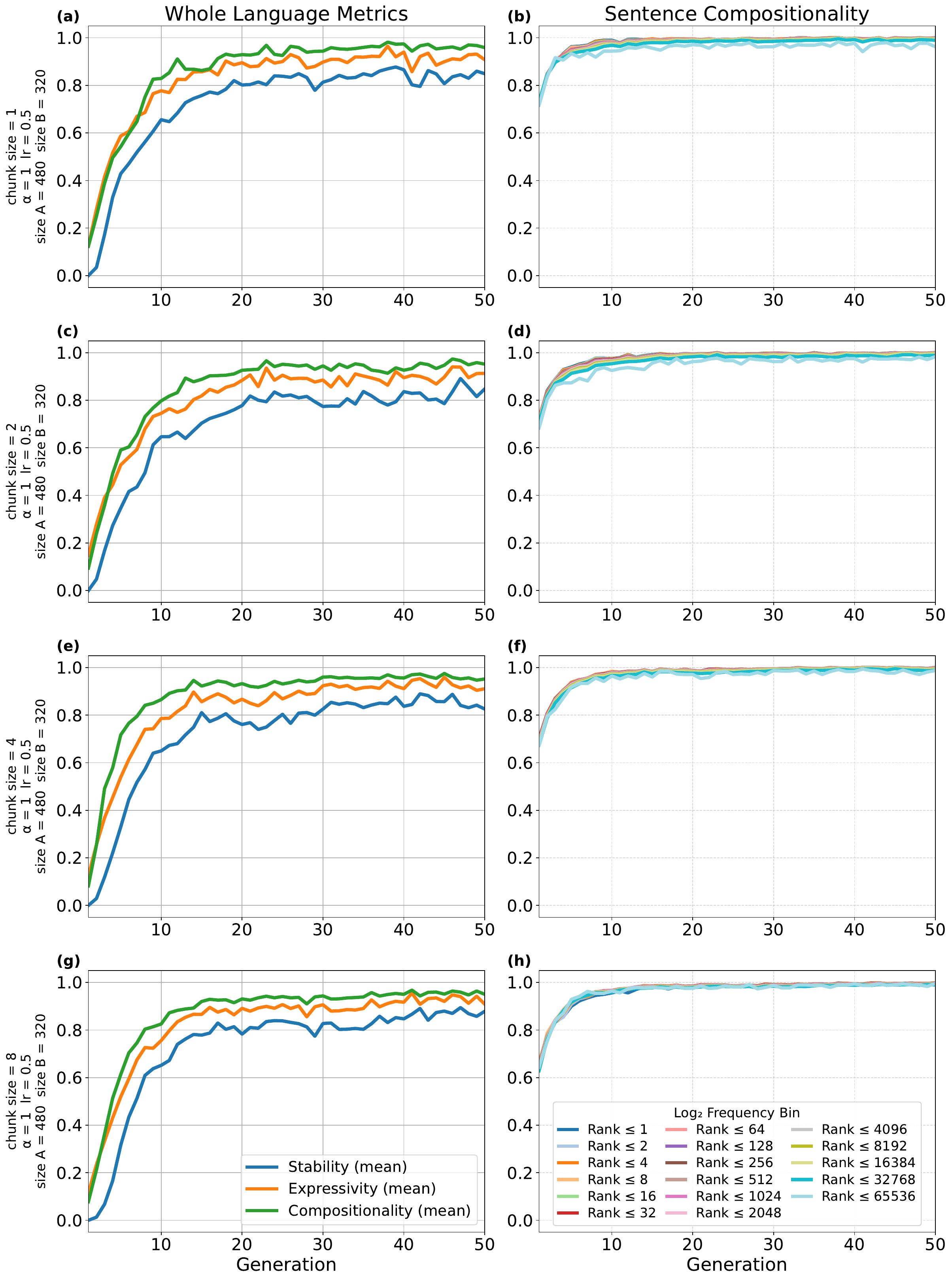}
\caption{Distribution skewed at the chunk level. Left: expressivity, compositionality, and stability for different chunk sizes. Right: compositionality as a function of $\log_2$ frequency-rank bins for each chunk size. Meanings are grouped, at the level of complete meaning vectors, into logarithmic frequency-rank bins: bin~0 contains the highest-frequency meaning, bin~1 the next highest-frequency group, and so on.}
\label{appendix02_all_chunks_comparison}
\end{figure}

\end{document}